\documentclass[sigconf]{acmart} 
\renewcommand\footnotetextcopyrightpermission[1]{} 
\usepackage{times}  
\usepackage{xcolor}
\usepackage[utf8]{inputenc}
\usepackage{caption}
\usepackage{url}  
\usepackage{graphicx}  
\usepackage{subcaption}
\usepackage{amsmath}
\usepackage{amsfonts}
\usepackage{bbm}
\usepackage[ruled,linesnumbered]{algorithm2e}
\usepackage{algpseudocode}
\usepackage{multirow}

\begin{document}
\title{One-Class Adversarial Nets for Fraud Detection}


 \author{Panpan Zheng}
 \affiliation{\institution{University of Arkansas}}
 \email{pzheng@uark.edu}

 \author{Shuhan Yuan}
 \affiliation{\institution{University of Arkansas}}
 \email{sy005@uark.edu}

 \author{Xintao Wu}
 \affiliation{\institution{University of Arkansas}}
 \email{xintaowu@uark.edu}

 \author{Jun Li}
 \affiliation{\institution{University of Oregon}}
 \email{lijun@cs.uoregon.edu}

 \author{Aidong Lu}
 \affiliation{\institution{\mbox{University of North Carolina at Charlotte}}}
 \email{aidong.lu@uncc.edu}

\acmConference[CIKM’18]{CIKM 2018 International Conference on Information and Knowledge Management}{October 22--26 2018}{Turin, Italy}
\acmYear{2018}

\begin{abstract}
Many online applications, such as online social networks or knowledge bases, are often attacked by malicious users who commit different types of actions such as vandalism on Wikipedia or fraudulent reviews on eBay. Currently, most of the fraud detection approaches require a training dataset that contains records of both benign and malicious users. However, in practice, there are often no or very few records of malicious users. In this paper, we develop one-class adversarial nets (OCAN) for fraud detection with only benign users as training data. OCAN first uses LSTM-Autoencoder to learn the representations of benign users from their sequences of online activities.  It then detects malicious users by training a discriminator of a complementary GAN model that is different from the regular GAN model. Experimental results show that our OCAN outperforms the state-of-the-art one-class classification models and achieves comparable performance with the latest multi-source LSTM model that requires both benign and malicious users in the training phase.

\end{abstract}

\keywords{fraud detection, one-class classification, generative adversarial nets}

\maketitle

\section{Introduction}

Online platforms such as online social networks (OSNs) and knowledge bases play a major role in online communication and knowledge sharing. However, there are various malicious users who conduct various fraudulent actions, such as spams, rumors, and vandalism, imposing severe security threats to OSNs and their legitimate participants. For example, the crowdsourcing mechanism of Wikipedia attracts the attention of vandals who involve in various vandalism actions like spreading false or misleading information to Wikipedia users. Meanwhile, fraudsters in the OSNs can also easily register fake accounts, inject fake content, or take fraudulent activities. To protect legitimate users, most Web platforms have tools or mechanisms to block malicious users. For example, Wikipedia adopts ClueBot NG \cite{Contributors2010Cluebot} to detect and revert obvious bad edits, thus helping administrators to identify and block vandals.

Detecting malicious users has also attracted increasing attention in the research community \cite{Cheng2017Anyone,Kumar2017Army,Yuan2017SpectrumBased,Kumar2015Vews,Ying2011Spectrum}. For example, research in \cite{Kumar2015Vews} focused on predicting whether a Wikipedia user is a vandal based on his edits. The VEWS system adopted a set of behavior features based on user edit-patterns, and used several traditional classifiers (e.g., random forest or SVM) to detect vandals.
To improve detection accuracy and avoid manual feature reconstruction, a multi-source long-short term memory network (M-LSTM) was proposed to detect vandals \cite{Yuan2017Wikipedia}. M-LSTM is able to capture different aspects of user edits and the learned user representations can be further used to analyze user behaviors. However, these detection models are trained over a training dataset that consists of both positive data (benign users) and negative data (malicious users). In practice, there are often no or very few records from malicious users in the collected training data. Manually labeling a large number of malicious users is tedious.

In this work, we tackle the problem of identifying malicious users when only benign users are observed. The basic idea is to adopt a generative model to generate malicious users with only given benign users. Generative adversarial networks (GAN) as generative models have demonstrated impressive performance in modeling the real data distribution and generating high quality synthetic data that is similar to real data \cite{Goodfellow2014Generative,Radford2015Unsupervised}. However, given benign users, a regular GAN model is unable to generate malicious users.



We develop one-class adversarial nets (OCAN) for fraud detection. During training, OCAN contains two phases. First, OCAN adopts the LSTM-Autoencoder \cite{Srivastava2015Unsupervised} to encode the benign users into a hidden space based on their online activities, and the encoded vectors are called benign user representations. Then, OCAN trains improved generative adversarial nets in which the discriminator is trained to be a classifier for distinguishing benign users and malicious users with the generator producing potential malicious users. To this end, we adopt the idea that the generator is trained to generate complementary samples instead of matching the original data distribution \cite{Dai2017Good}. In particular, we propose a complementary GAN model. The generator of the complementary GAN aims to generate samples that are complementary to the representations of benign users, i.e., the potential malicious users. The discriminator is trained to separate benign users and complementary samples. Since the behaviors of malicious users and that of benign users are complementary, we expect the discriminator can distinguish benign users and malicious users. By combining the encoder of LSTM-Autoencoder and the discriminator of the complementary GAN, OCAN can accurately predict whether a new user is benign or malicious based on his online activities.

The advantages of OCAN for fraud detection are as follows. First, since OCAN does not require any information about malicious users, we do not need to manually compose a mixed training dataset, thus more adaptive to different types of malicious user identification tasks.
Second, different from existing one-class classification models, OCAN generates complementary samples of benign users and trains the discriminator to separate complementary samples from benign users, enabling the trained discriminator to better separate malicious users from benign users. Third, OCAN can capture the sequential information of user activities. After training, the detection model can adaptively update a user representation once the user commits a new action and predict whether the user is a fraud or not dynamically.

\section{Related Work}

{\noindent \bf Fraud detection:}
Due to the openness and anonymity of Internet, the online platforms attract a large number of malicious users, such as vandals, trolls, and sockpuppets. Many fraud detection techniques have been developed in recent years \cite{Akoglu2015Graph,Jiang2014Catchsync,Cao2014Uncovering,Ying2011Spectrum,Kumar2018False}, including content-based approaches and graph-based approaches. The content-based approaches extract content features, (i.e., text, URL), to identify malicious users from user activities on social networks \cite{Benevenuto2010Detecting}. Meanwhile, graph-based approaches identify frauds based on network topologies. Research in \cite{Yuan2017SpectrumBased} proposed two deep neural networks for fraud detection on a signed graph. Often based on unsupervised learning, the graph-based approaches consider fraud as anomalies and extract various graph features associated with nodes, edges, ego-net, or communities from the graph \cite{Akoglu2015Graph,Noble2003GraphBased,Manzoor2016Fast}.

Fraud detection is also related to the malicious behavior and misinformation detection, including detecting the vandalism edits on Wikipedia or Wikidata, rumor and fake review detection. Research in \cite{Heindorf2016Vandalism} developed both content and context features of a Wikidata revision to identify the vandalism edit. Research in \cite{Kumar2016Disinformation} focused on detecting hoaxes on Wikipedia by finding characteristic in terms of article content and features of the editor who created the hoax. In~\cite{Mukherjee2013What}, different types of behavior features were extracted and used to detect fake reviews on Yelp. Research in \cite{Lim2010Detecting} have identified several representative behaviors of review spammers. Research in \cite{Rayana2015Collective} proposed a framework that combined the text, metadata as well as relational data to detect suspicious users and reviews. Research in \cite{Xie2012Review} studied the co-anomaly patterns in multiple review-based time series. Some researches further focused on detecing frauders who delibrately evaded the detection by mimicing normal users \cite{Wang2017Gang,Hooi2016Fraudar}.

{\noindent \bf Deep neural network:}
Deep neural networks have achieved promising results in computer vision, natural language processing, and speech recognition \cite{Lecun2015Deep}. Recurrent neural network as one type of deep neural networks is widely used for modeling time sequence data \cite{Graves2013Generating,Neubig2017Neural}. However, it is difficult to train standard RNNs over long sequences of text because of gradient vanishing and exploding \cite{Bengio1997Learning}.  Long shot-term Memory (LSTM) \cite{Hochreiter1997Long} was proposed to model temporal sequences and capture their long-range dependencies more accurately than the standard RNNs. LSTM-Autoencoder is a sequence-to-sequence model that has been widely used for paragraph generation \cite{Li2015Hierarchical,Nallapati2016Abstractive,Sutskever2014Sequence}, video representation \cite{Srivastava2015Unsupervised}, etc. GAN is a framework for estimating generative models via an adversarial process \cite{Goodfellow2014Generative}. Recently, generative adversarial nets (GAN) have achieved great success in computer vision tasks, including image generation \cite{Radford2015Unsupervised,Springenberg2016Unsupervised,Ledig2017PhotoRealistic} and image classification \cite{Chen2016Infogan,Springenberg2016Unsupervised,Odena2016Conditional}. Currently,  the GAN model is usually applied on two or multi-class datasets instead of one-class.

{\noindent \bf One-class classification:}
One-class classification (OCC) algorithms aim to build classification models when only one class of samples are observed and the other class of samples are absent \cite{Khan2014OneClass}, which is also related to the novelty detection \cite{Pimentel2014Review}. One-class support vector machine (OSVM), as one of widely adopted for one class classification, aims to separate one class of samples from all the others by constructing a hyper-sphere around the observed data samples \cite{Tax2004Support,Manevitz2001OneClass}. Other traditional classification models also extend to the one-class scenario. For example, one-class nearest neighbor (OCNN) \cite{Tax2001Uniform} predicts the class of a sample based on its distance to its nearest neighbor in the training dataset. One-class Gaussian process (OCGP) chooses a proper GP prior and derives membership scores for one-class classification \cite{Kemmler2013OneClass}. However, OCNN and OCGP need to set a threshold to detect another class of data. The threshold is either set by a domain expert or tuned based on a small set of two-class labeled data. In this work, we propose a framework that combines LSTM-Autoencoder and GAN to detect vandals with only knowing benign users. To our best knowledge, this is the first work that examines the use of deep learning models for fraud detection when only one-class training data is available. Meanwhile, comparing to existing one-class algorithms, our model trains a classifier by generating a large number of ``novel'' data and does not require any labeled data to tune parameters.

\section{Preliminary}
\subsection{Long Short-Term Memory Network}
Long short-term memory network (LSTM) is one type of recurrent neural network.
Given a sequence $\mathcal{X}=(\mathbf{x}_1, \dots, \mathbf{x}_t, \dots, \mathbf{x}_T)$ where $\mathbf{x}_t \in \mathbb{R}^{d}$ denotes the input at the $t$-th step, LSTM maintains a hidden state vector $\mathbf{h}_t \in \mathbb{R}^{h}$ to keep track the sequence information from the current input $\mathbf{x}_t$ and the previous hidden state $\mathbf{h}_{t-1}$.
The hidden state $\mathbf{h}_t$ is computed by
\begin{align}
	\label{eq:lstm}
	\begin{split}
	\mathbf{\tilde{c}}_t &= \tanh (\mathbf{W}_{c} \mathbf{x}_t + \mathbf{U}_{c} \mathbf{h}_{t-1} + \mathbf{b}_c), \\
	\mathbf{i}_t &= \sigma (\mathbf{W}_{i} \mathbf{x}_t + \mathbf{U}_{i} \mathbf{h}_{t-1} + \mathbf{b}_i), \\
	\mathbf{f}_t &= \sigma (\mathbf{W}_{f} \mathbf{x}_t + \mathbf{U}_{f} \mathbf{h}_{t-1} + \mathbf{b}_f), \\
	\mathbf{o}_t &= \sigma (\mathbf{W}_{o} \mathbf{x}_t + \mathbf{U}_{o} \mathbf{h}_{t-1} + \mathbf{b}_o), \\
	\mathbf{c}_t &= \mathbf{i}_t \odot \mathbf{\tilde{c}}_t + \mathbf{f}_t \odot \mathbf{c}_{t-1}, \\
	\mathbf{h}_t &= \mathbf{o}_t \odot \tanh(\mathbf{c}_t), \\
	\end{split}
\end{align}
where $\sigma$ is the sigmoid function; $\odot$ represents element-wise product; $\mathbf{i}_t$, $\mathbf{f}_t$, $\mathbf{o}_t$, $\mathbf{c}_t$ indicate the input gate, forget gate, output gate, and cell activation vectors and $\mathbf{\tilde{c}}_t$ denotes the intermediate vector of cell state; $\mathbf{W}$ and $\mathbf{U}$ are the weight parameters; $\mathbf{b}$ is the bias term.

We simplify the update of each LSTM step described in Equation \ref{eq:lstm} as
\begin{equation}
	\mathbf{h}_{t} = LSTM(\mathbf{x}_{t}, \mathbf{h}_{t-1}),
\end{equation}
where $\mathbf{x}_{t}$ is the input of the current step; $\mathbf{h}_{t-1}$ is the hidden vector of the last step; $\mathbf{h}_{t}$ indicates the output of the current step.

\subsection{Generative Adversarial Nets}

Generative adversarial nets (GAN) are generative models that consist of two components: a generator $G$ and a discriminator $D$. Typically, both $G$ and $D$ are multilayer neural networks. $G(\mathbf{z})$ generates fake samples from a prior $p_{\mathbf{z}}$ on a noise variable $\mathbf{z}$ and learns a generative distribution $p_G$ to match the real data distribution $p_{\text{data}}$. On the contrary, the discriminative model $D$ is a binary classifier that predicts whether an input is a real data $\mathbf{x}$ or a generated fake data from $G(\mathbf{z})$. Hence, the objective function of $D$ is defined as:
\begin{equation}
\label{eq:reg_d_loss}
\max\limits_D \quad \mathbb{E}_{\mathbf{x} \sim p_{\text{data}}}[\log D(\mathbf{x})] + \mathbb{E}_{\mathbf{z} \sim p_{\mathbf{z}}}[\log(1-D(G(\mathbf{z})))],
\end{equation}
where $D(\cdot)$ outputs the probability that $\cdot$ is from the real data rather than the generated fake data. In order to make the generative distribution $p_G$ close to the real data distribution $p_{\text{data}}$, $G$ is trained by fooling the discriminator not be able to distinguish the generated data from the real data. Thus, the objective function of $G$ is defined as:
\begin{equation}
\label{eq:reg_g_loss}
\min\limits_G \quad \mathbb{E}_{\mathbf{z} \sim p_{\mathbf{z}}}[\log(1-D(G(\mathbf{z})))].
\end{equation}
Minimizing the Equation \ref{eq:reg_g_loss} is achieved if the discriminator is fooled by generated data $G(\mathbf{z})$ and predicts high probability that $G(\mathbf{z})$ is real data.

Overall, GAN is formalized as a minimax game $\min\limits_G \max\limits_D V(G,D)$ with the value function:
\begin{equation}
V(G,D)= \mathbb{E}_{\mathbf{x} \sim p_{\text{data}}}[\log D(\mathbf{x})] + \mathbb{E}_{\mathbf{z} \sim p_{\mathbf{z}}}[\log(1-D(G(\mathbf{z})))].
\end{equation}

Theoretical analysis shows that GAN aims to minimize the Jensen-Shannon divergence between the data distribution $p_{\text{data}}$ and the generative distribution $p_G$ \cite{Goodfellow2014Generative}. The minimization of JS divergence is achieved when $p_G = p_{\text{data}}$. Therefore, GAN is trained by distinguishing the real data and generated fake data.


\begin{figure*}[htb]
  \centering
	  \includegraphics[width=0.8\textwidth]{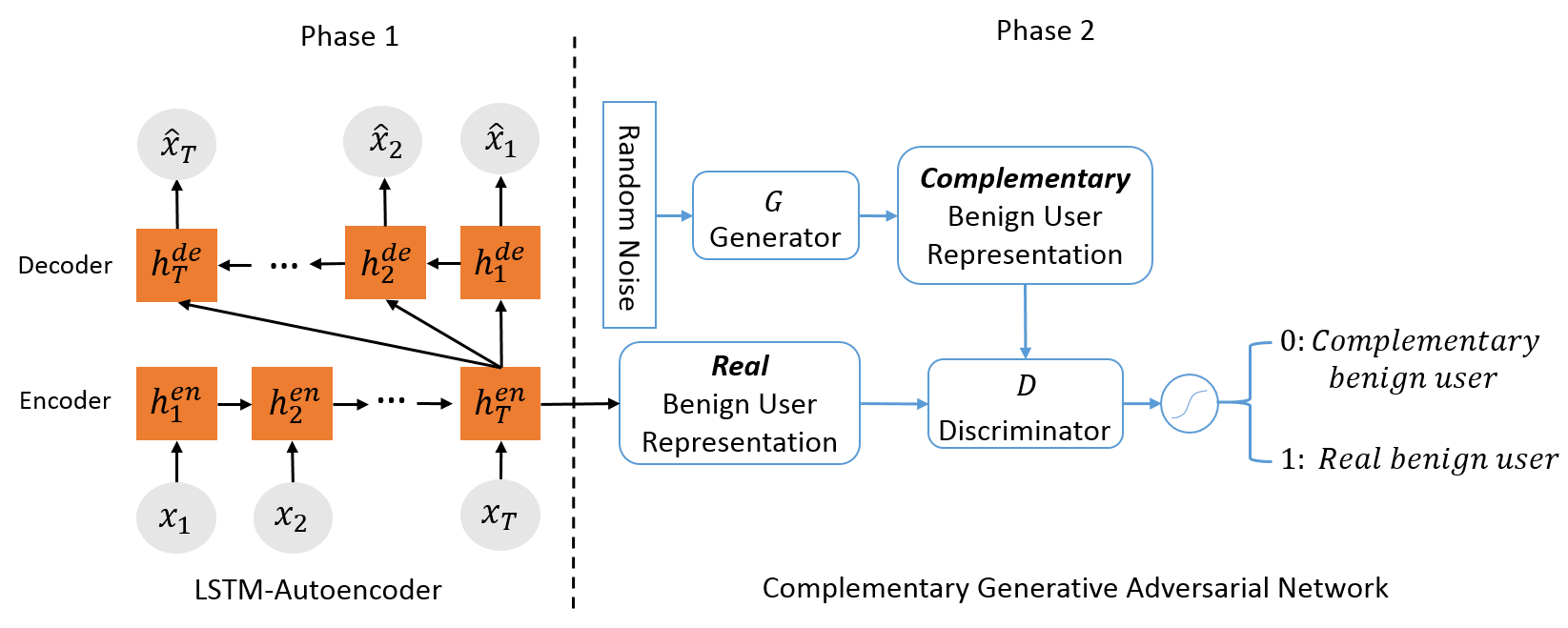}
	  \caption{The training framework of OCAN}
	  \label{fig:train}
\end{figure*}

\section{OCAN: One-Class Adversarial Nets}

\subsection{Framework Overview}


OCAN contains two phases during training. The first phase is to learn user representations. As shown in the left side of Figure \ref{fig:train}, LSTM-Autoencoder is adopted to learn the benign user representations from the benign user activity sequences. The LSTM-Autoencoder model is a sequence-to-sequence model that consists of two LSTM models as the encoder and decoder respectively. The encoder computes the hidden representation of an input, and the decoder computes the reconstructed input based on the hidden representation. The trained LSTM-Autoencoder can capture the salient information of users' activity sequences because the objective function is to make the reconstructed input close to the original input.
Furthermore, the encoder of the trained LSTM-Autoencoder, when deployed for fraud detection, is expected to map the benign users and malicious users to relatively separate regions in the continuous feature space because the activity sequences of benign and malicious users are different.


Given the user representations, the second phase is to train a complementary GAN with a discriminator that can clearly distinguish the benign and malicious users. The generator of the complementary GAN aims to generate complementary samples that are in the low-density area of benign users, and the discriminator aims to separate the real and complementary benign users. The discriminator then has the ability to detect malicious users which locate in separate regions from benign users. The framework of training complementary GAN for fraud detection is shown in the right side of Figure \ref{fig:train}.


The pseudo-code of training OCAN is shown in Algorithm \ref{algr:train}. Given a training dataset $M_{\text{benign}}$ that contains activity sequence feature vectors of $N$ benign users, we first train the LSTM-Autoencoder model (Lines \ref{algr:start_ae}--\ref{algr:end_ae}). After training the LSTM-Autoencoder, we adopt the encoder in the LSTM-Autoencoder model to compute the benign user representation (Lines \ref{algr:start_u}--\ref{algr:end_u}). Finally, we use the benign user representation to train the complementary GAN (Lines \ref{algr:start_gan}--\ref{algr:end_gan}). For simplicity, we write the algorithm with a minibatch size of 1, i.e., iterating each user in the training dataset to train LSTM-Autoencoder and GAN. In practice, we sample $m$ real benign users and use the generator to generate $m$ complementary samples in a minibatch. In our experiments, the size of minibatch is 32.

Our OCAN moves beyond the naive approach of adopting a regular GAN model in the second phase. The generator of a regular GAN aims to generate the representations of fake benign users that are close to the representations of real benign users. The discriminator of a regular GAN is to identify whether an input is a representation of a real benign user or a fake benign user from the generator. However, one potential drawback of the regular GAN is that once the discriminator is converged, the discriminator cannot have high confidence on separating real benign users from real malicious users. We denote the OCAN with the regular GAN as OCAN-r and compare its performance with OCAN in the experiment.

\begin{algorithm}[htb]
	\SetAlFnt{\small}
	\DontPrintSemicolon
	\SetKwInOut{Inputs}{Inputs}\SetKwInOut{Outputs}{Outputs}
	\Inputs{Training dataset $M_{\text{benign}}=\{\mathcal{X}_{1}, \cdots, \mathcal{X}_{N}\}$,  \\
			Training epochs for LSTM-Autoencoder \\  $Epoch_{AE}$ and GAN $Epoch_{GAN}$}
	\Outputs{Well-trained LSTM-Autoencoder and complementary GAN}

	initialize parameters in LSTM-Autoencoder and complementary GAN;

	$j \leftarrow 0$;

	\While{$j<Epoch_{AE}$}{
	\label{algr:start_ae}
		\ForEach {user $u$ in $M_{\text{benign}}$}{			
			compute the reconstructed sequence of user activities by LSTM-Autoencoder (Eq. \ref{eq:en}, \ref{eq:de}, and \ref{eq:re});
			\label{line:decoder}
			
			optimize the parameters in LSTM-Autoencoder with the loss function Eq. \ref{eq:ae_loss};
			\label{line:ae_optimize}
		}

		$j \leftarrow j+1$;
	}
	\label{algr:end_ae}

	$\mathcal{V}=\emptyset$;

	\ForEach {user $u$ in $M_{\text{benign}}$}{
	\label{algr:start_u}
		compute the benign user representation $\mathbf{v}_u$ by the encoder of LSTM-Autoencoder (Eq. \ref{eq:en}, \ref{eq:user});

		$\mathcal{V} += \mathbf{v}_u$;
	}
	\label{algr:end_u}

	$j \leftarrow 0$;

	\While{$j<Epoch_{GAN}$}{
	\label{algr:start_gan}
		\ForEach {benign user representation $\mathbf{v}_u$ in $\mathcal{V}$}{
			optimize the discriminator $D$ and generator $G$ with loss functions Eq. \ref{eq:d_loss}, \ref{eq:g_loss}, respectively;
		}
	}
	\label{algr:end_gan}
	\Return well-trained LSTM-Autoencoder and complementary GAN
\caption{Training One-Class Adversarial Nets}
\label{algr:train}
\end{algorithm}



\subsection{LSTM-Autoencoder for User Representation}

The first phase of OCAN is to encode users to a continuous hidden space. Since each online user has a sequence of activities (e.g., edit a sequence of pages), we adopt LSTM-Autoencoder to transform a variable-length user activity sequence into a fixed-dimension user representation. Formally, given a user $u$ with $T$ activities, we represent the activity sequence as $\mathcal{X}_u=(\mathbf{x}_1, \dots, \mathbf{x}_t, \dots, \mathbf{x}_T)$ where $\mathbf{x}_t \in \mathbb{R}^{d}$ is the $t$-th activity feature vector.


\textbf{Encoder:} The encoder encodes the user activity sequence $\mathcal{X}_u$ to a user representation with an LSTM model:
\begin{equation}
\label{eq:en}
	\mathbf{h}_{t}^{en} = LSTM^{en}(\mathbf{x}_{t}, \mathbf{h}_{t-1}^{en}),
\end{equation}
where $\mathbf{x}_{t}$ is the feature vector of the $t$-th activity; $\mathbf{h}_{t}^{en}$ indicates the $t$-th hidden vector of the encoder.

The last hidden vector $\mathbf{h}_{T}^{en}$ captures the information of a whole user activity sequence and is considered as the user representation $\mathbf{v}$:
\begin{equation}
\label{eq:user}
	\mathbf{v}=\mathbf{h}_{T}^{en}.
\end{equation}

\textbf{Decoder:} In our model, the decoder adopts the user representation $\mathbf{v}$ as the input to reconstruct the original user activity sequence $\mathcal{X}$:
\begin{align}
	\label{eq:de}
	\mathbf{h}_{t}^{de} &= LSTM^{de}(\mathbf{v}, \mathbf{h}_{t-1}^{de}), \\
	\label{eq:re}
	\hat{\mathbf{x}}_{t} &= f(\mathbf{h}_{t}^{de}),
\end{align}
where $\mathbf{h}_{t}^{de}$ is the $t$-th hidden vector of the decoder; $\hat{\mathbf{x}}_{t}$ indicates the $t$-th reconstructed activity feature vector; $f(\cdot)$ denotes a neural network to compute the sequence outputs from hidden vectors of the decoder. Note that we adopt $\mathbf{v}$ as input of the whole sequence of the decoder, which has achieved great performance on sequence-to-sequence models \cite{Cho2014Learning}.

The objective function of LSTM-Autoencoder is:
\begin{equation}
\label{eq:ae_loss}
	\mathcal{L}_{(AE)}(\hat{\mathbf{x}}_{t}, \mathbf{x}_{t}) = \sum_{t=1}^T{(\hat{\mathbf{x}}_{t} - \mathbf{x}_{t})^2},
\end{equation}
where $\mathbf{x}_{t}$ ($\hat{\mathbf{x}}_{t}$) is the $t$-th (reconstructed) activity feature vector. After training, the last hidden vector of encoder $\mathbf{h}_T$ can reconstruct the sequence of user feature vectors. Thus, the  representation of user $\mathbf{v}=\mathbf{h}_{T}^{en}$ captures the salient information of user behavior.


\subsection{Complementary GAN}

The generator $G$ of complementary GAN is a feedforward neural network where its output layer has the same dimension as the user representation $\mathbf{v}$. Formally, we define the generated samples as $\tilde{\mathbf{v}}= G(\mathbf{z})$. Unlike the generator in a regular GAN which is trained to match the distribution of the generated fake benign user representation with that of benign user representation $p_{\text{data}}$, the generator $G$ of complementary GAN learns a generative distribution $p_G$ that is close to the complementary distribution $p^*$ of the benign user representations, i.e., $p_G = p^*$.
The complementary distribution $p^*$ is defined as:
\begin{equation}
\label{eq:complement}
	p^*(\tilde{\mathbf{v}}) =
		\begin{cases}
			 \frac{1}{\tau} \frac{1}{p_{\text{data}}(\tilde{\mathbf{v}})} \quad &\text{if } p_{\text{data}}(\tilde{\mathbf{v}}) > \epsilon \text{ and } \tilde{\mathbf{v}} \in \mathcal{B}_\mathbf{v} \\
			 C \quad &\text{if } p_{\text{data}}(\tilde{\mathbf{v}}) \leq \epsilon \text{ and } \tilde{\mathbf{v}} \in \mathcal{B}_\mathbf{v},
		\end{cases}
\end{equation}
where $\epsilon$ is a threshold to indicate whether the generated samples are in high-density regions; $\tau$ is a normalization term; $C$ is a small constant; $\mathcal{B}_\mathbf{v}$ is the space of user representation. To make the generative distribution $p_G$ close to the complementary distribution $p^*$, the complementary generator $G$ is trained to minimize the KL divergence between $p_G$ and $p^*$. Based on the definition of KL divergence, we have the following objective function:
\begin{equation}
\label{eq:kl}
\begin{split}
	\mathcal{L}_{KL(p_G \parallel p^*)} &= -\mathcal{H}(p_{G}) - \mathbb{E}_{\tilde{\mathbf{v}} \sim p_{G}}\log p^*(\tilde{\mathbf{v}}) \\
			   &= -\mathcal{H}(p_{G}) + \mathbb{E}_{\tilde{\mathbf{v}} \sim p_{G}}\log p_{\text{data}}(\tilde{\mathbf{v}}) \mathbbm{1}[p_{\text{data}}(\mathbf{\tilde{\mathbf{v}}})>\epsilon] \\
			   & + \mathbb{E}_{\tilde{\mathbf{v}} \sim p_{G}}(\mathbbm{1}[p_{\text{data}}(\mathbf{\tilde{\mathbf{v}}})>\epsilon]\log\tau - \mathbbm{1}[p_{\text{data}}(\mathbf{\tilde{\mathbf{v}}}) \leq \epsilon]\log C),
\end{split}
\end{equation}
where $\mathcal{H}(\cdot)$ is the entropy, and $\mathbbm{1}[\cdot]$ is the indicator function. The last term of Equation \ref{eq:kl} can be omitted because both $\tau$ and $C$ are constant terms and the gradients of the indicator function $\mathbbm{1}[\cdot]$ with respect to parameters of the generator are mostly zero.

Meanwhile, the complementary generator $G$ adopts the feature matching loss \cite{Salimans2016Improved} to ensure that the generated samples are constrained in the space of user representation $\mathcal{B}_\mathbf{v}$.
\begin{equation}
\label{eq:fm}
	\mathcal{L}_{\text{fm}} =  \parallel \mathbb{E}_{\tilde{\mathbf{v}} \sim p_{G}}  f(\tilde{\mathbf{v}}) - \mathbb{E}_{\mathbf{v} \sim p_{\text{data}}} f(\mathbf{v})  \parallel^2,
\end{equation}
where $f(\cdot)$ denotes the output of an intermediate layer of the discriminator used as a feature representation of $\mathbf{v}$.

Thus, the complete objective function of the generator is defined as:
\begin{equation}
\label{eq:g_loss}
\begin{split}
	\min\limits_G \quad &-\mathcal{H}(p_{G})+\mathbb{E}_{\tilde{\mathbf{v}} \sim p_{G}}\log p_{\text{data}}(\tilde{\mathbf{v}}) \mathbbm{1}[p_{\text{data}}(\tilde{\mathbf{v}})>\epsilon] \\
		  &+ \parallel \mathbb{E}_{\tilde{\mathbf{v}} \sim p_{G}}  f(\tilde{\mathbf{v}}) - \mathbb{E}_{\mathbf{v} \sim p_{\text{data}}} f(\mathbf{v})  \parallel^2.
\end{split}
\end{equation}
Overall, the objective function of the complementary generator aims to let the generative distribution $p_G$ close to the complementary samples $p^*$, i.e., $p_G=p^*$, and make the generated samples from different regions (but in the same space of user representations) than those of the benign users.

\begin{figure}[h]
  \centering
  	\begin{subfigure}[t]{0.23\textwidth}
		\centering
		\includegraphics[width=0.8\textwidth,height=1.2in]{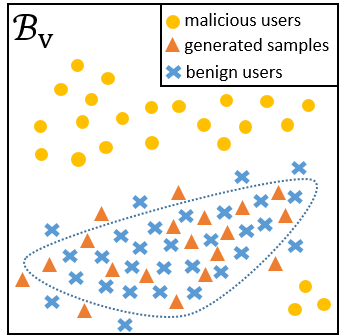}
		\caption{Regular GAN}
		\label{fig:regular_ocan}
	\end{subfigure}
	~
	\begin{subfigure}[t]{0.23\textwidth}
		\centering
		\includegraphics[width=0.8\textwidth,height=1.2in]{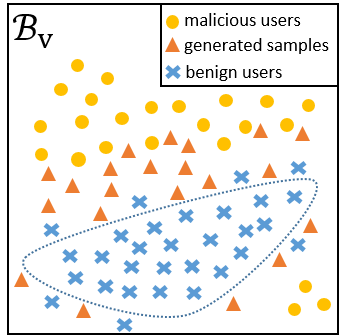}
		\caption{Complementary GAN}
		\label{fig:one_class_ocan}
	\end{subfigure}
	  \caption{Demonstrations of the ideal generators of regular GAN and complementary GAN. The blue dot line indicates the high density   regions of benign users.}
	  \label{fig:illustration}
\end{figure}

Figure \ref{fig:illustration} illustrates the difference of the generators of regular GAN and complementary GAN. The objective function of the generator of regular GAN in Equation \ref{eq:reg_g_loss} is trained to fool the discriminator by generating fake benign users similar to the real benign users. Hence, as shown in Figure \ref{fig:regular_ocan}, the generator of regular GAN generates the distribution of fake benign users that have the similar distribution of real benign users in the feature space. On the contrary, the objective function of the generator of complementary GAN in Equation \ref{eq:g_loss} is trained to generate complementary samples that are in the low-density regions of benign users (shown in Figure \ref{fig:one_class_ocan}).

To optimize the objective function of generator, we need to approximate the entropy of generated samples $\mathcal{H}(p_{G})$ and the probability distribution of real samples $p_{\text{data}}$. To minimize $-\mathcal{H}(p_{G})$, we adopt the pull-away term (PT) proposed by \cite{Zhao2016EnergyBased,Dai2017Good} that encourages the generated feature vectors to be orthogonal. The PT term increases the diversity of generated samples and can be considered as a proxy for minimizing $-\mathcal{H}(p_{G})$. The PT term is defined as
\begin{equation}
\label{eq:pt}
\mathcal{L}_{PT}=\frac{1}{N(N-1)}\sum_{i}^N\sum_{j \neq i}^N (\frac{f(\tilde{\mathbf{v}}_i)^T f(\tilde{\mathbf{v}}_j)}{\parallel f(\tilde{\mathbf{v}}_i)\parallel \parallel f(\tilde{\mathbf{v}}_j)\parallel})^2,
\end{equation}
where $N$ is the size of a mini-batch.
The probability distribution of real samples $p_{\text{data}}$ is usually unavailable, and approximating $p_{\text{data}}$ is computationally expensive. We adopt the approach proposed by \cite{Schoneveld2017SemiSupervised} that a discriminator from a regular GAN can detect whether the data from the real data distribution $p_{\text{data}}$ or from the generator's distribution. The basic idea is that the discriminator is able to detect whether a sample is from the real data distribution $p_{\text{data}}$ or from the generator when the generator is trained to generate samples that are close to real benign users. Hence, the discriminator is sufficient to identify the data points that are above a threshold of $p_{\text{data}}$ during training. We separately train a regular GAN model based on benign user representations and use the discriminator of the regular GAN as a proxy to evaluate $p_{\text{data}}(\tilde{\mathbf{v}}) > \epsilon$.


The discriminator $D$ takes the benign user representation $\mathbf{v}$ and generated user representation $\tilde{\mathbf{v}}$ as inputs and tries to distinguish $\mathbf{v}$ from $\tilde{\mathbf{v}}$. As a classifier, $D$ is a standard feedforward neural network with a softmax function as its output layer, and the objective function of $D$ is:
\begin{equation}
\label{eq:d_loss}
\begin{split}
	\max\limits_D \quad & \mathbb{E}_{\mathbf{v} \sim p_{\text{data}}} [\log D(\mathbf{v})] + \mathbb{E}_{\tilde{\mathbf{v}} \sim p_{G}}[\log (1-D(\tilde{\mathbf{v}}))] + \\ & \mathbb{E}_{\mathbf{v} \sim p_{\text{data}}} [D(\mathbf{v})\log D(\mathbf{v})].
\end{split}
\end{equation}
The first two terms in Equation \ref{eq:d_loss} are the objective function of discriminator in the regular GAN model. Therefore, the discriminator of complementary GAN is trained to separate the benign users and complementary samples. The last term in Equation \ref{eq:d_loss} is a conditional entropy term which encourages the discriminator to detect real benign users with high confidence. Then, the discriminator is able to separate the benign and malicious users clearly.

Although the objective functions of the discriminators of regular GAN and complementary GAN are similar, the capabilities of discriminators of regular GAN and complementary GAN for malicious detection are different. The discriminator of regular GAN aims to separate the benign users and generated fake benign users. However, after training, the generated fake benign users locate in the same regions as the real benign users (shown in Figure \ref{fig:regular_ocan}). The probabilities of real and generated fake benign users predicted by the discriminator of regular GAN are all close to 0.5. Thus, giving a benign user, the discriminator cannot predict the benign user with high confidence. On the contrary, the discriminator of complementary GAN is trained to separate the benign users and generated complementary samples. Since the generated complementary samples have the same distribution as the malicious users (shown in Figure \ref{fig:one_class_ocan}), the discriminator of complementary GAN can also detect the malicious users.

\begin{figure}[htb]
    \centering
  	\includegraphics[height=4cm, keepaspectratio]{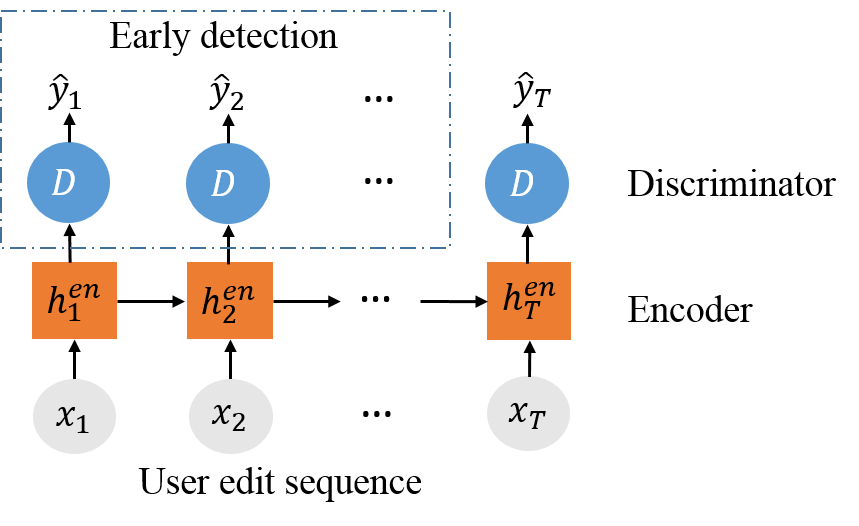}
  	\caption{The fraud detection model}
  	\label{fig:test}
\end{figure}

\begin{algorithm}[h]
	\DontPrintSemicolon
	\SetKwInOut{Inputs}{Inputs}\SetKwInOut{Outputs}{Outputs}
	\Inputs{Testing dataset $M_{\text{test}}=\{\mathcal{X}_{1}, \cdots, \mathcal{X}_{N}\}$,  \\
			Well-trained LSTM-Autoencoder and GAN}
	\Outputs{the user labels $\mathcal{Y}$ in $M_{\text{test}}$ }

	$\mathcal{Y}=\emptyset$;

	\ForEach {user $u$ in $M_{\text{test}}$}{

		compute the user representation $\mathbf{v}_u$ by the encoder in LSTM-Autoencoder (Eq. \ref{eq:en}, \ref{eq:user});
		\label{algr:ur}

		predict the label $\hat y_u$ of the user by $D(\mathbf{v}_u)$
		\label{algr:pr}

		$\mathcal{Y} += \hat y_u$
	}

	\Return the user labels $\mathcal{Y}$
\caption{Fraud Detection}
\label{algr:test}
\end{algorithm}

\section{Fraud Detection Model}
Although the training procedure of OCAN contains two phases that train LSTM-Autoencoder and complementary GAN successively, the fraud detection model is an end-to-end model.
We show the procedure pseudocode of detecting fraud in Algorithm \ref{algr:test} and illustrate its structure in Figure \ref{fig:test}. To detect a malicious user, we first compute the user representation $\mathbf{v}_u$ based on the encoder in the LSTM-Autoencoder model (Line \ref{algr:ur}). Then, we predict the user label based on the discriminator of complementary GAN, i.e., $p(\hat {y}_u|\mathbf{v}_u) = D(\mathbf{v}_u)$.


{\textbf{Early fraud detection:}}
The upper-left region of Figure \ref{fig:test} shows that our OCAN model can also achieve early detection of malicious users. Given a user $u$, at each step $t$, the hidden states $\mathbf{h}_{u_t}^{en}$ are updated until the $t$-th step by taking the current feature vector $\mathbf{x}_{u_t}$ as input and are able to capture the user behavior information until the $t$-th step. Thus, the user representation at the $t$-th step is denoted as $\mathbf{v}_{u_t} = \mathbf{h}_{u_t}^{en}$. Finally, we can use the discriminator $D$ to calculate the probability $p(\hat {y}_{u_t}|\mathbf{v}_{u_t}) = D(\mathbf{v}_{u_t})$ of the user to be a malicious user based on the current step user representation $\mathbf{v}_t$.


\section{Experiments}
\begin{table*}[htb]
\centering
\caption{Vandal detection results (mean$\pm$std.) on precision, recall, F1 and accuracy}
\label{tb:results}
\begin{tabular}{|c|c|c|c|c|c|}
\hline
Input                                & Algorithm     & Precision             & Recall              & F1                  & Accuracy            \\ \hline
\multirow{3}{*}{Raw feature vector}  & OCNN          & $0.5680 \pm 0.0129$   & $0.8646 \pm 0.0599$ & $0.6845 \pm 0.0184$ & $0.6027 \pm 0.0161$  \\ \cline{2-6}
                                     & OCGP          & $0.5767 \pm 0.0087$   & $0.9000 \pm 0.0560$ & $0.7023 \pm 0.0193$ & $0.6196 \pm 0.0142$ \\ \cline{2-6}
                                     & OCSVM         & $0.6631 \pm 0.0057$   & $0.9829 \pm 0.0011$ & $0.7919 \pm 0.0040$ & $0.7417 \pm 0.0064$ \\ \hline \hline
\multirow{4}{*}{User representation} & OCNN          & $0.8314 \pm 0.0351$   & $0.8028 \pm 0.0476$ & $0.8150 \pm 0.0163$ & $0.8181 \pm 0.0153$ \\ \cline{2-6}
                                     & OCGP          & $0.8381 \pm,0.0225$   & $0.8289 \pm 0.0374$ & $0.8326 \pm 0.0158$ & $0.8337 \pm 0.0139$ \\ \cline{2-6}
                                     & OCSVM         & $0.6558 \pm 0.0058$   & \textbf{$\mathbf{0.9590 \pm 0.0096}$} & $0.7789 \pm 0.0064$ & $0.7278 \pm 0.0080$ \\ \cline{2-6}
                                     & OCAN          & \textbf{$\mathbf{0.9067 \pm 0.0615}$}   & $0.9292 \pm 0.0348$ & \textbf{$\mathbf{0.9010 \pm 0.0228}$} & \textbf{$\mathbf{0.8973 \pm 0.0244}$} \\ \hline \hline
User representation                  & OCAN-r          & $0.8673 \pm 0.0355$   & $0.8759 \pm 0.0529$ & $0.8701 \pm 0.0267$ & $0.8697 \pm 0.0244$ \\
                                      \hline
\end{tabular}
\end{table*}

\subsection{Experiment Setup}

{\noindent \bf Dataset:}
To evaluate OCAN, we focus on one type of malicious users, i.e., vandals on Wikipedia. We conduct our evaluation on UMDWikipedia dataset \cite{Kumar2015Vews}. This dataset contains information of around 770K edits from Jan 2013 to July 2014 (19 months) with 17105 vandals and 17105 benign users. Each user edits a sequence of Wikipedia pages. We keep those users with the lengths of edit sequence range from 4 to 50. After this preprocessing, the dataset contains 10528 benign users and 11495 vandals.

To compose the feature vector $\mathbf{x}_t$ of the user's $t$-th edit, we adopt the following edit features: (1) whether or not the user edited on a meta-page; (2) whether or not the user consecutively edited the pages less than 1 minutes; (3) whether or not the user's current edit page had been edited before; (4) whether or not the user's current edit would be reverted.

We further evaluate our model on a credit card transaction dataset in Section \ref{sec:credit}. Although it is not a sequence dataset, it can still be used to compare the performance of OCAN against baselines in the context of one-class fraud detection.

{\noindent \bf Hyperparameters:}
For LSTM-Autoencoder, the dimension of the hidden layer is 200, and the training epoch is 20. For the complementary GAN model, both discriminator and generator are feedforward neural networks. Specifically, the discriminator contains 2 hidden layers which are 100 and 50 dimensions. The generator takes the 50 dimensions of noise as input, and there is one hidden layer with 100 dimensions. The output layer of the generator has the same dimension as the user representation which is 200 in our experiments. The training epoch of complementary GAN is 50. The threshold $\epsilon$ defined in Equation \ref{eq:g_loss} is set as the 5-quantile probability of real benign users predicted by a pre-trained discriminator. We evaluated several values from 4-quantile to 10-quantile and found the results are not sensitive.

{\noindent \bf Repeatability:} Our software together with the datasets used in this paper are available at https://github.com/PanpanZheng/OCAN

\subsection{Comparison with One-Class Classification}

{\noindent \bf Baselines:} We compare OCAN with the following widely used one-class classification approaches:
\begin{itemize}
	\item One-class nearest neighbors (\textbf{OCNN}) \cite{Tax2001Uniform} labels a testing sample based on the distance from the sample to its nearest neighbors in training dataset and the average distance of those nearest neighbors. If the difference between these two distances is larger than a threshold, the testing sample is an anomaly.
	\item One-class Gaussian process (\textbf{OCGP}) \cite{Kemmler2013OneClass} is a one-class classification model based on Gaussian process regression.
	\item One-class SVM (\textbf{OCSVM}) \cite{Tax2004Support} adopts support vector machine to learn a decision hypersphere around the positive data, and considers samples located outside this hypersphere as anomalies.
\end{itemize}

For baslines, we use the implementation provided in NDtool~\footnote{\url{http://www.robots.ox.ac.uk/~davidc/publications_NDtool.php}}. The hyperparameters of baselines set as default values in NDtool.
Note that both OCNN and OCGP require a small portion (5\% in our experiments) of vandals as a validation dataset to tune an appropriate threshold for vandal detection. However, OCAN does not require any vandals for training and validation. Since the baselines are not sequence models, we compare OCAN to baselines in two ways. First, we concatenate all the edit feature vectors of a user to a \textit{raw feature vector} as an input to baselines. Second, the baselines have the same inputs as the discriminator, i.e., the \textit{user representation} $\mathbf{v}$ computed from the encoder of LSTM-Autoencoder. Meanwhile, OCAN cannot adopt the raw feature vectors as inputs to detect vandals. This is because GAN is only suitable for real-valued data \cite{Goodfellow2014Generative}.


To evaluate the performance of vandal detection, we randomly select 7000 benign users as the training dataset and 3000 benign users and 3000 vandals as the testing dataset. We report the mean value and standard deviation based on 10 different runs. Table \ref{tb:results} shows the means and standard deviations of the precision, recall, F1 score and accuracy for vandal detection.  First, OCAN achieves better performances than baselines in terms of F1 score and accuracy in both input settings. It means the discriminator of complementary GAN can be used as a one-class classifier for vandal detection.  We can further observe that when the baselines adopt the raw feature vector instead of user representation, the performances of both OCNN and OCGP decrease significantly. It indicates that the user representations computed by the encoder of LSTM-Autoencoder capture the salient information about user behavior and can improve the performance of one-class classifiers. However, we also notice that the standard deviations of OCAN are higher than the baselines with user representations as inputs. We argue that this is because GAN is widely known for difficult to train. Thus, the stability of OCAN is relatively lower than the baselines.

Furthermore, we show the experimental results of OCAN-r, which adopts the regular GAN model instead of the complementary GAN in the second training phase of OCAN, in the last row of Table \ref{tb:results}. We can observe that the performance of OCAN is better than OCAN-r. It indicates that the discriminator of complementary GAN which is trained on real and complementary samples can more accurately separate the benign users and vandals.


\begin{table}[htb]
\centering
\caption{Early Vandal detection results on precision, recall, F1, and the average number of edits before the vandals are blocked}
\label{tb:early}
\begin{small}
\begin{tabular}{|c|c|c|c|c|c|}
\hline
       & Vandals & Precision & Recall & F1       & Edits \\ \hline
\multirow{5}{*}{M-LSTM}
       & 7000          & 0.8416    & 0.9637 & 0.8985   & 7.21      \\  \cline{2-6}
       & 1000          & 0.9189    & 0.8910 & 0.9047   & 5.98      \\ \cline{2-6}
       & 400           & 0.9639    & 0.6767 & 0.7951   & 3.64      \\ \cline{2-6}
       & 300           & 0.0000    & 0.0000 & 0.0000   & 0.00      \\ \cline{2-6}
       & 100           & 0.0000    & 0.0000 & 0.0000   & 0.00      \\
\hline
OCAN   & 0			   & \textbf{0.8014}  & \textbf{0.9081} & \textbf{0.8459}   & \textbf{7.23} \\ \hline \hline
OCAN-r  & 0			   & 0.7228      		& 0.8968     		& 0.7874      		  & 7.18 \\ \hline
\end{tabular}
\end{small}
\end{table}

\subsection{Comparison with M-LSTM for Early Vandal Detection}

We further compare the performance of OCAN in terms of early vandal detection with one latest deep learning based vandal detection model, M-LSTM, developed in \cite{Yuan2017Wikipedia}. Note that M-LSTM assumes a training dataset that contains both vandals and benign users. In our experiments, we train our OCAN with the training data consisting of 7000 benign users and no vandals and train M-LSTM with a training data consisting the same 7000 benign users and a varying number of vandals (from 7000 to 100). For OCAN and M-LSTM, we use the same testing dataset that contains 3000 benign users and 3000 vandals. Note that in OCAN and M-LSTM, the hidden state $\mathbf{h}_t^{en}$ of the LSTM model captures the up-to-date user behavior information and hence we can achieve early vandal detection. The difference is that the M-LSTM model uses $\mathbf{h}_t^{en}$ as the input of a classifier directly whereas OCAN further trains complementary GAN and uses its discriminator as a classifier to make the early vandal detection. In this experiment, instead of applying the classifier on the final user representation $\mathbf{v}=\mathbf{h}_T^{en}$, the classifiers of M-LSTM and OCAN are applied on each step of LSTM hidden state $\mathbf{h}_t^{en}$ and predict whether a user is a vandal after the user commits the t-th action.

Table \ref{tb:early} shows comparison results in terms of the precision, recall, F1 of early vandal detection, and the average number of edits before the vandals were truly blocked.  We can observe that OCAN achieves a comparable performance as the M-LSTM when the number of vandals in the training dataset is large (1000, 4000, and 7000). However, M-LSTM has very poor accuracy when the number of vandals in the training dataset is small. In fact, we observe that M-LSTM could not detect any vandal when the training dataset contains less than 400 vandals. On the contrary, OCAN does not need any vandal in the training data.

The experimental results of OCAN-r for early vandal detection are shown in the last row of Table \ref{tb:early}. OCAN-r outperforms M-LSTM when M-LSTM is trained on a small number of the training dataset. However, the OCAN-r is not as good as OCAN. It indicates that generating complementary samples to train the discriminator can improve the performance of the discriminator for vandal detection.

\subsection{OCAN Framework Analysis}

\begin{figure}[htb]
  \centering
	  \includegraphics[width=0.45\textwidth]{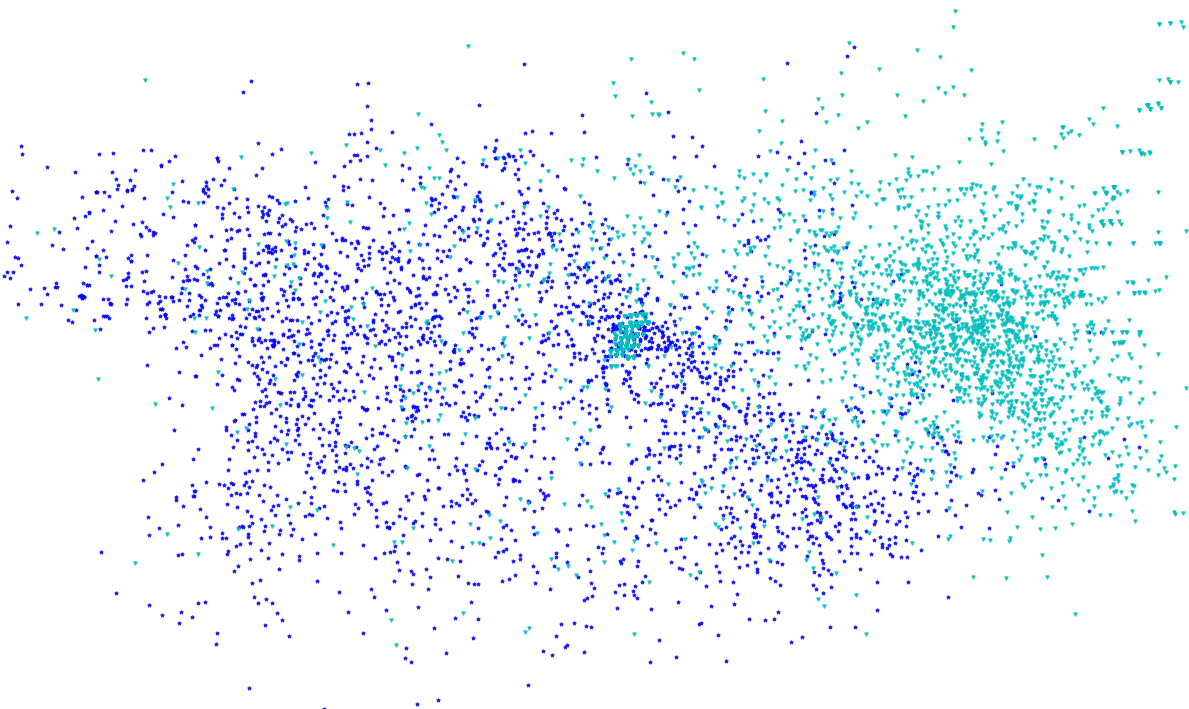}
  \caption{Visualization of 3000 benign users (blue star) and 3000 vandals (cyan triangle) based on user representation.}
  \label{fig:user}
\end{figure}

{\noindent \bf LSTM-Autoencoder:}
The reason of adopting LSTM-Autoencoder is that it transforms edit sequences to user representation. It also helps encode benign users and vandals to relatively different locations in the hidden space, although the LSTM-Autoencoder is only trained by benign users. To validate our intuition, we obtain user representations of the testing dataset by the encoder in the LSTM-Autoencoder. Then, we map those user representations to a two-dimensional space based on the Isomap approach \cite{Tenenbaum2000Global}. Figure \ref{fig:user} shows the visualization of user representations. We observe that the benign users and vandals are relatively separated in the two-dimensional space, indicating the capability of LSTM-Autoencoder.

\begin{figure*}[htb]
	\begin{subfigure}[t]{0.25\textwidth}
		\centering
		\includegraphics[height=1.4in]{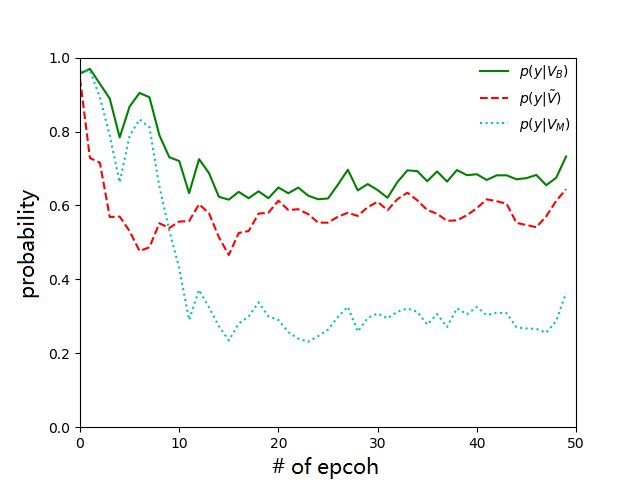}
		\caption{Prob. predicted by OCAN}
		\label{fig:probability_ocan}
	\end{subfigure}
	~
	\begin{subfigure}[t]{0.25\textwidth}
		\centering
		\includegraphics[height=1.4in]{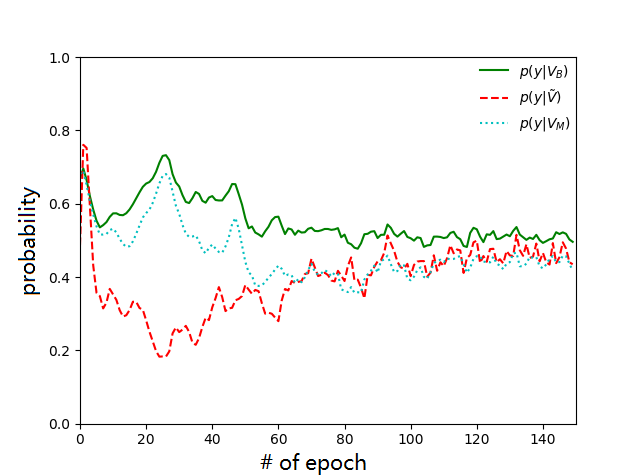}
		\caption{Prob. predicted by OCAN-r}
		\label{fig:probability_ocan-}
	\end{subfigure}
	~
	\begin{subfigure}[t]{0.24\textwidth}
		\centering
		\includegraphics[height=1.4in]{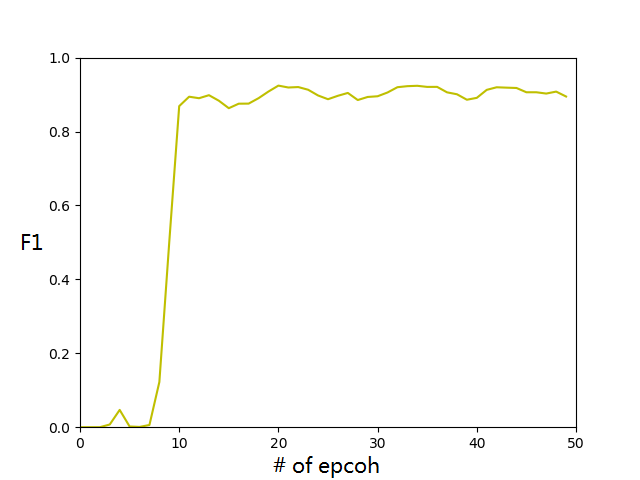}
		\caption{F1 score of OCAN}
		\label{fig:f1_ocan}
	\end{subfigure}
	~
	\begin{subfigure}[t]{0.24\textwidth}
		\centering
		\includegraphics[height=1.4in]{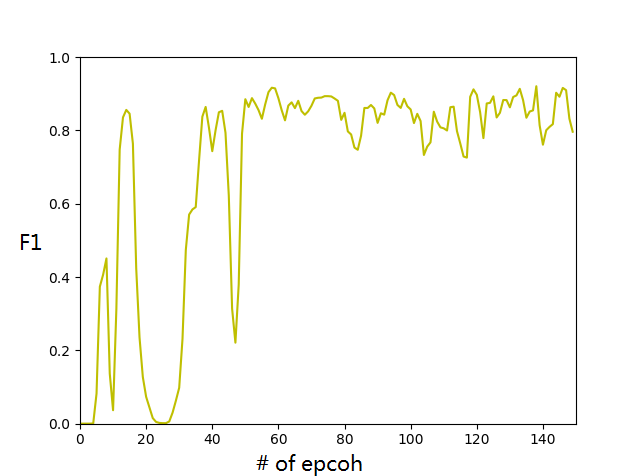}
		\caption{F1 score of OCAN-r}
		\label{fig:f1_ocan-}
	\end{subfigure}

	\caption{Training progresses of OCAN (\ref{fig:probability_ocan},\ref{fig:f1_ocan}) and OCAN-r(\ref{fig:probability_ocan-},\ref{fig:f1_ocan-}). Three lines in Figures \ref{fig:probability_ocan} and \ref{fig:probability_ocan-} indicate the probabilities of being benign users predicted by the discriminator: real benign users $p(y|\mathbf{v}_B)$ (green line) vs. generated samples $p(y|\tilde{\mathbf{v}})$ (red broken line) vs. real malicious users $p(y|\mathbf{v}_M)$ (blue dotted line). Figures \ref{fig:f1_ocan} and \ref{fig:f1_ocan-} show the F1 scores of OCAN and OCAN-r during training.}
	\label{fig:train_progress}
\end{figure*}

{\noindent \bf Complementary GAN vs. Regular GAN:}

In our OCAN model, the generator of complementary GAN aims to generate complementary samples that lie in the low-density region of real samples, and the discriminator is trained to detect the real and complementary samples.
We examine the training progress of OCAN in terms of predication accuracy. We calculate probabilities of real benign users $p(y|\mathbf{v}_B)$ (shown as green line in Figure \ref{fig:probability_ocan}), malicious users $p(y|\mathbf{v}_M)$ (blue dotted line) and generated samples $p(y|\tilde{\mathbf{v}})$ (read broken line) being benign users predicted by the discriminator of complementary GAN on the testing dataset after each training epoch. We can observe that after OCAN is converged, the probabilities of malicious users predicted by the discriminator of complementary GAN are much lower than that of benign users. For example, at the epoch 40, the average probability of real benign users $p(y|\mathbf{v}_B)$ predicted by OCAN is around $70\%$, while the average probability of malicious users $p(y|\mathbf{v}_M)$ is only around $30\%$. Meanwhile, the average probability of generated complementary samples $p(y|\tilde{\mathbf{v}})$ lies between the probabilities of benign and malicious users.

On the contrary, the generator of a regular GAN in the OCAN-r model aims to generate fake samples that are close to real samples, and the discriminator of GAN focuses on distinguishing the real and generated fake samples. As shown in Figure \ref{fig:probability_ocan-}, the probabilities of real benign users and probabilities of malicious users predicted by the discriminator of regular GAN become close to each other during training. After the OCAN-r is converged, say epoch 120, both the probabilities of real benign users and malicious users are close to 0.5. Meanwhile, the probability of generated samples is similar to the probabilities of real benign users and malicious users.


We also show the F1 scores of OCAN and OCAN-r on the testing dataset after each training epoch in Figure \ref{fig:f1_ocan} and \ref{fig:f1_ocan-}. We can observe that the F1 score of OCAN-r is not as stable as (and also a bit lower than) OCAN. This is because the outputs of the discriminator for real and fake samples are close to 0.5 after the regular GAN is converged. If the probabilities of real benign users predicted by the discriminator of the regular GAN swing around 0.5, the accuracy of vandal detection will fluctuate accordingly.


We can observe from Figure \ref{fig:train_progress} another nice property of OCAN compared with OCAN-r for fraud detection, i.e., OCAN is converged faster than OCAN-r. We can observe that OCAN is converged with only training 20 epochs while the OCAN-r requires nearly 100 epochs to keep stable. This is because the complementary GAN is trained to separate the benign and malicious users while the regular GAN mainly aims to generate fake samples that match the real samples. In general, matching two distributions requires more training epochs than separating two distributions. Meanwhile, the feature matching term adopted in the generator of complementary GAN is also able to improve the training process \cite{Salimans2016Improved}.



\begin{figure}[htb]
  \centering
	  \includegraphics[width=0.45\textwidth, height=1.5in]{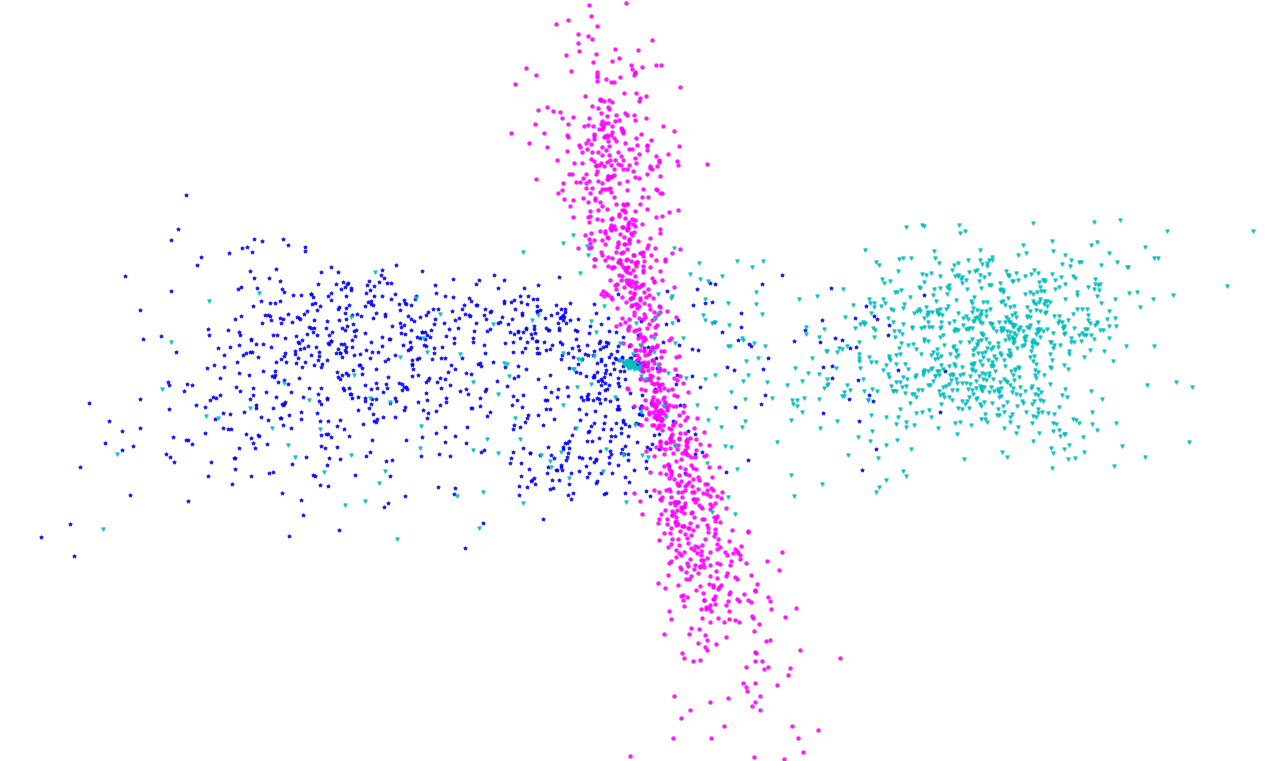}
  \caption{2D visualization of three types of users: real benign (blue star), vandal (cyan triangle), and complementary benign (red dot)}
  \label{fig:2d_bgv}
\end{figure}

{\noindent \bf Visualization of three types of users:}
We project the user representations of the three types of users (i.e., benign, vandal and complementary benign generated by OCAN) to a two-dimensional space by Isomap. Figure \ref{fig:2d_bgv} visualizes the three types of users. We observe that the generated complementary users lie in the low-density regions of real benign users. Meanwhile, the generated samples are also between the benign users and vandals. Since the discriminator is trained to separate the benign and complementary benign users, the discriminator is able to separate benign users and vandals.


\begin{table}[htb]
\centering
\caption{Clustering results of DBSCAN}
\label{tb:dbscan_cluster}
\begin{tabular}{|c|c|c|c|c|}
\hline
   & Cluster Size  & Benign & Vandal  & Complement  \\
\hline
C1       & 2537      & 2448           & 89        & 0     \\
\hline
C2        & 2420     & 93            & 2327         & 0     \\
\hline
C3        & 2840     & 0             & 0      & 2840     \\
\hline
Isolated   & 1203      & 459           & 584       & 160     \\
\hline

\end{tabular}
\end{table}

\begin{table*}[]
\centering
\caption{Fraud detection results (mean$\pm$std.) on precision, recall, F1 and accuracy of credit card fraud detection}
\label{tb:cc_resutls}
\begin{tabular}{|c|c|c|c|c|c|}
\hline
Input                                & Algorithm & Precision                    & Recall                       & F1                           & Accuracy                     \\ \hline
\multirow{4}{*}{Raw feature vector}  & OCNN      & $0.8029 \pm 0.1497$          & $0.9075 \pm 0.0596$          & $0.8383 \pm 0.0660$          & $0.8132 \pm 0.1082$          \\ \cline{2-6}
                                     & OCGP      & $0.9700 \pm 0.0222$          & $0.7308 \pm 0.0812$          & $0.8302 \pm 0.0450$          & $0.8531 \pm 0.0324$          \\ \cline{2-6}
                                     & OCSVM     & $0.6590 \pm 0.0100$          & \textbf{$\mathbf{0.9404 \pm 0.0017}$} & $0.7749 \pm 0.0068$          & $0.7267 \pm 0.0107$          \\ \cline{2-6}
                                     & OCAN      & \textbf{$\mathbf{0.9755 \pm 0.0110}$} & $0.7416 \pm 0.0498$          & \textbf{$\mathbf{0.8416 \pm 0.0330}$} & \textbf{$\mathbf{0.8613 \pm 0.0243}$} \\ \hline
\multirow{4}{*}{Transaction representation} & OCNN      & $0.7058 \pm 0.1396$          & $0.9390 \pm 0.0786$          & $0.7910 \pm 0.0608$          & $0.7401 \pm 0.1079$          \\ \cline{2-6}
                                     & OCGP      & $0.8813 \pm 0.1177$          & $0.8566 \pm 0.0822$          & $0.8576 \pm 0.0417$          & $0.8536 \pm 0.0623$          \\ \cline{2-6}
                                     & OCSVM     & $0.6547 \pm 0.0151$          & \textbf{$\mathbf{0.9509 \pm 0.0101}$} & $0.7755 \pm 0.0127$          & $0.7245 \pm 0.0185$          \\ \cline{2-6}
                                     & OCAN      & \textbf{$\mathbf{0.9067 \pm 0.0614}$} & $0.8320 \pm 0.0319$          & \textbf{$\mathbf{0.8656 \pm 0.0220}$} & \textbf{$\mathbf{0.8701 \pm 0.0264}$} \\ \hline
\end{tabular}
\end{table*}

{\noindent \bf User clustering:}
To further analyze the complementary GAN model, we adopt the classic DBSCAN algorithm \cite{Ester1996DensityBased} to cluster 3000 benign users, 3000 vandals from the testing dataset, and 3000 generated complementary benign users. Table \ref{tb:dbscan_cluster} shows clustering results including cluster size and class distributions of each cluster. We set the maximum radius of the neighborhood $\epsilon=1.4305$ (the average distances among the user representations) and the minimum number of points $minPts=180$. We observe three clusters where C1 is benign users, C2 is vandal, and C3 is complementary samples in addition to 1203 isolated users that could not form any cluster. We emphasize that our OCAN can still make good predictions of those isolated points accurately (with 89\% accuracy).

We further calculate the centroid of C1, C2 and C3 based on their user representations and adopt the centroids to calculate distances among each type of users. The distance between the centroids of real benign users and complementary benign users is 3.6346, while the distance between the centroids of real benign users and vandals is 3.888. Since the discriminator is trained to identify real benign users and complementary benign users, the discriminator can detect vandals which have larger distances to real benign users than that of complementary benign users.


\subsection{Case Study on Credit Card Fraud Detection}
\label{sec:credit}
We further evaluate our model on a credit card fraud detection dataset \footnote{\url{https://www.kaggle.com/dalpozz/creditcardfraud}}. The dataset records credit card transactions in two days and has 492 frauds out of 284,807 transactions. Each transaction contains 28 features. We adopt 700 genuine transactions as a training dataset and 490 fraud and 490 genuine transactions as a testing dataset. Since the transaction features of the dataset are numerical data derived from PCA, OCAN is able to detect frauds by using raw features as inputs. Meanwhile, we also evaluate the performance of OCAN in the hidden feature space. Because the transaction in credit card dataset is not a sequence data, we adopt the regular autoencoder model instead of LSTM-autoencoder to obtain the transaction representations. In our experiments, the dimension of transaction representation is 50.

Table \ref{tb:cc_resutls} shows the classification results of credit card fraud detection. Overall, the performance of OCAN and baselines are similar to the results of vandal detection shown in Table \ref{tb:results}. OCAN achieves the best accuracy and F1 with both input settings. Meanwhile, the performance of OCAN using transaction representations as inputs is better than using raw features. It shows that OCAN can outperform the existing one-class classifiers in different datasets and can be applied to detect different types of malicious users.

\begin{figure}[h]
    \centering
	\includegraphics[width=0.38\textwidth]{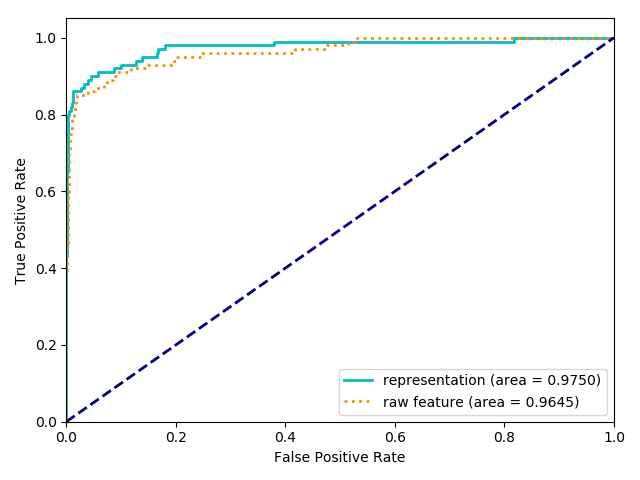}
	\caption{ROC curve of OCAN on an imbalanced dataset}
	\label{fig:auc}
\end{figure}

{\bf \noindent Testing on an imbalanced dataset.} In the real scenario, there are more genuine transactions than fraud transactions. Hence, after training OCAN on 700 genuine transactions, we further test OCAN on an imbalanced dataset which consists of 1000 genuine transactions and 100 fraud transactions. Figure \ref{fig:auc} shows the ROC curve of OCAN on the unbalanced dataset. It indicates OCAN achieves promising performance for fraud detection with adopting raw feature vector and transaction representation as inputs on the imbalanced dataset. In particular, the AUC of OCAN with raw features is 0.9645, and the AUC of OCAN with transaction representation is 0.9750.

\section{Conclusion}
In this paper, we have developed OCAN that consists of LSTM-Autoencoder and complementary GAN for fraud detection when only benign users are observed during the training phase. During training, OCAN adopts LSTM-Autoencoder to learn benign user representations, and then uses the benign user representations to train a complementary GAN model. The generator of complementary GAN can generate complementary benign user representations that are in the low-density regions of real benign user representations, while the discriminator is trained to distinguish the real and complementary benign users. After training, the discriminator is able to detect malicious users which are outside the regions of benign users. We have conducted theoretical and empirical analysis to demonstrate the advantages of complementary GAN over regular GAN.
We conducted experiments over two real world datasets and showed OCAN outperforms the state-of-the-art one-class classification models. Moreover, our OCAN model can also achieve early vandal detection since OCAN takes the user edit sequences as inputs and achieved comparable accuracy with the latest M-LSTM model that needs both benign and vandal users in the training data. In our future work, we plan to extend the techniques for fraud detection in the semi-supervised learning scenario.

\begin{acks}
The authors acknowledge the support from National Science Foundation to Panpan Zheng and Xintao Wu (1564250),  Jun Li (1564348) and Aidong Lu (1564039).
\end{acks}



\bibliography{Remote}
\bibliographystyle{ACM-Reference-Format}
\end{document}